\documentclass[10pt,twocolumn,letterpaper]{article}

\usepackage[pagenumbers]{cvpr} % To force page numbers, e.g. for an arXiv version

\newcommand\blfootnote[1]{%
  \begingroup
  \renewcommand\thefootnote{}\footnote{#1}%
  \addtocounter{footnote}{-1}%
  \endgroup
}

\definecolor{cvprblue}{rgb}{0.21,0.49,0.74}
\usepackage[pagebackref,breaklinks,colorlinks,allcolors=cvprblue]{hyperref}

\usepackage{xcolor}         % colors
\usepackage{amsmath}
\usepackage{graphicx}
\usepackage{multirow}
\usepackage{subcaption}
\usepackage{placeins}
\usepackage{array}
\usepackage{calc}
\usepackage{bm}
\usepackage{capt-of}

\usepackage{xparse} % Safe modern way to patch commands

\let\origparagraph\paragraph
\renewcommand{\paragraph}[1]{\vspace{-2mm}\origparagraph{#1}}

\newcommand{\ours}{Radar2Shape}

\title{Radar2Shape: 3D Shape Reconstruction from High-Frequency Radar using Multiresolution Signed Distance Functions}

\author{
    Neel Sortur\textsuperscript{1} \quad
    Justin Goodwin\textsuperscript{2} \quad
    Purvik Patel\textsuperscript{1} \quad
    Luis Enrique Martinez Jr\textsuperscript{2} \\
    Tzofi Klinghoffer\textsuperscript{3} \quad
    Rajmonda S. Caceres\textsuperscript{2} \quad
    Robin Walters\textsuperscript{1}
    \vspace{2pt} \\ 
    \small{
    \textsuperscript{1}Northeastern University \quad
    \textsuperscript{2}MIT Lincoln Laboratory \quad
    \textsuperscript{3}Massachusetts Institute of Technology 
    }
    \vspace{2pt} \\ 
}

\newlength{\meshimageheight}

\begin{document}
\setlength{\meshimageheight}{0.125\textheight}

\twocolumn[{
\maketitle
\begin{center}
    \captionsetup{type=figure}
    \vspace{-6mm}
    \includegraphics[width=0.7\textwidth]{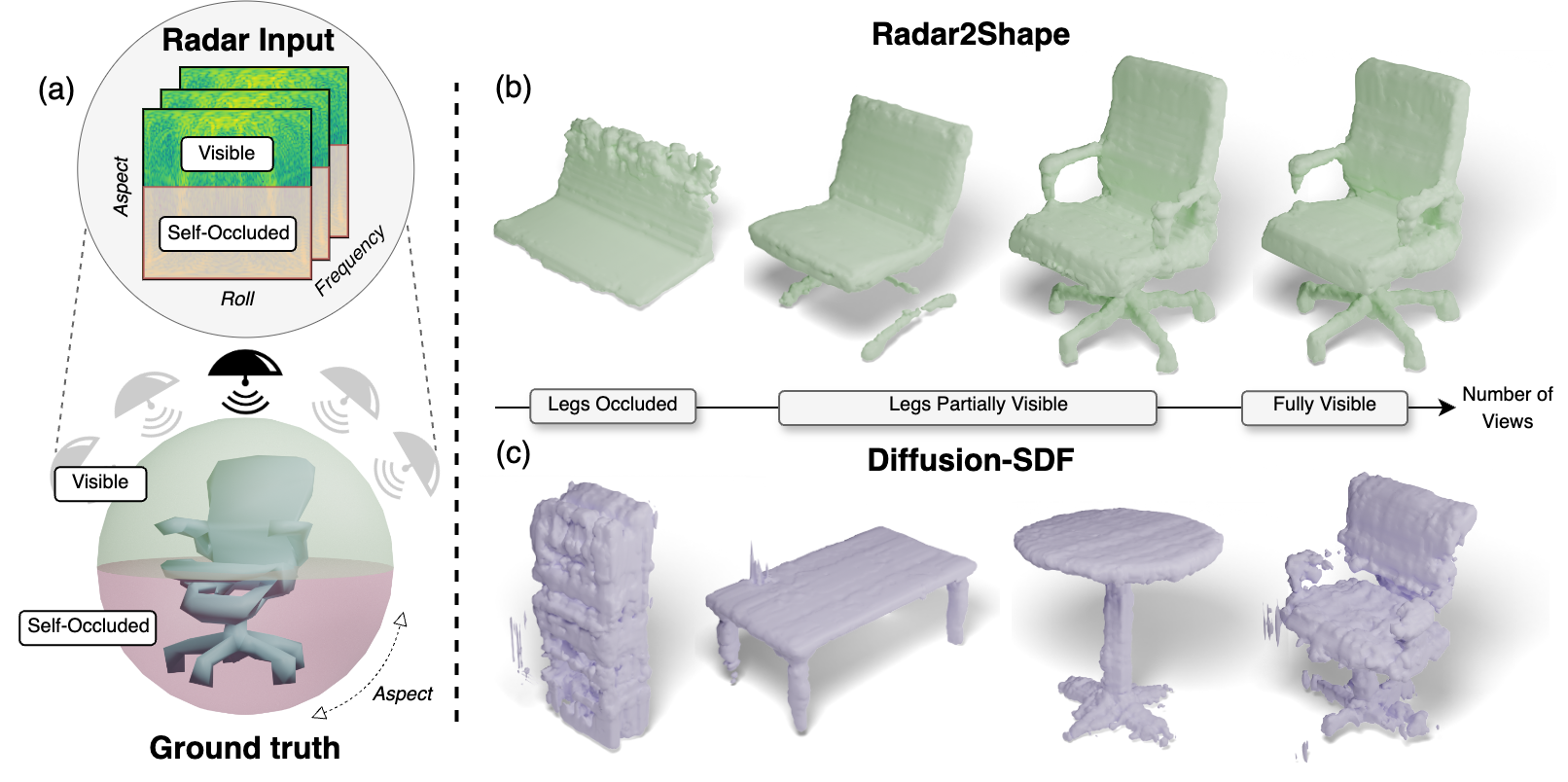}
    \vspace{-2mm}
    \captionof{figure}{\textbf{Overview.} \ours{} solves the challenging task of 3D shape reconstruction from radar captured at limiting viewing angles. \textbf{(a)} Limited views cause self-occlusion, resulting in missing information in the measurement. \textbf{(b)} Our approach overcomes this ambiguity by using a data-driven diffusion prior with a novel coarse-to-fine refinement technique in signed distance function space. This method accurately generates occluded geometries based on partial radar measurements, leading to better performance than \textbf{(c)} existing domain-adapted methods that can fail with limited views and struggle even in full observability.}
    \label{fig:figure1}
\end{center}
}]

\blfootnote{DISTRIBUTION STATEMENT A. Approved for public release. Distribution is unlimited.
This material is based upon work supported by the Under Secretary of War for Research and Engineering under Air Force Contract No. FA8702-15-D-0001 or FA8702-25-D-B002. Any opinions, findings, conclusions or recommendations expressed in this material are those of the author(s) and do not necessarily reflect the views of the Under Secretary of War for Research and Engineering.
© 2025 Massachusetts Institute of Technology.
Delivered to the U.S. Government with Unlimited Rights, as defined in DFARS Part 252.227-7013 or 7014 (Feb 2014). Notwithstanding any copyright notice, U.S. Government rights in this work are defined by DFARS 252.227-7013 or DFARS 252.227-7014 as detailed above. Use of this work other than as specifically authorized by the U.S. Government may violate any copyrights that exist in this work.}

\begin{abstract}
    Determining the shape of 3D objects from high-frequency radar signals is analytically complex but critical for commercial and aerospace applications. Previous deep learning methods have been applied to radar modeling; however, they often fail to represent arbitrary shapes or have difficulty with real-world radar signals which are collected over limited viewing angles. Existing methods in optical 3D reconstruction can generate arbitrary shapes from limited camera views, but struggle when they naively treat the radar signal as a camera view. In this work, we present \ours{}, a denoising diffusion model that handles a partially observable radar signal for 3D reconstruction by correlating its frequencies with multiresolution shape features. Our method consists of a two-stage approach: first, \ours{} learns a regularized latent space with hierarchical resolutions of shape features, and second, it diffuses into this latent space by conditioning on the frequencies of the radar signal in an analogous coarse-to-fine manner. We demonstrate that \ours{} can successfully reconstruct arbitrary 3D shapes even from partially-observed radar signals, and we show robust generalization to two different simulation methods and real-world data. Additionally, we release two synthetic benchmark datasets to encourage future research in the high-frequency radar domain so that models like \ours{} can safely be adapted into real-world radar systems.

\end{abstract}

\vspace{-3mm}

\section{Introduction}
\label{sec:intro}

Radar is a reliable sensing mechanism in adverse light and weather conditions with wide-ranging applications such as robotics~\cite{barnes2020}, autonomous driving~\cite{radar_autonomous_vehicles}, and remote sensing~\cite{BergenBGD02a}. It operates by transmitting radio waves and analyzing the response, echoes that return after striking an object. Types of radar are typically distinguished by their wavelength -- longer wavelengths struggle to detect small objects, like raindrops or geologic particles, while shorter wavelengths (high-frequency radar) can detect a variety of sizes, but may be more noisy. In both regimes, geometrically characterizing and reconstructing an object from its radar response still presents a challenging inverse learning task. At long ranges, radar signals are often noisy and provide poor resolution \cite{Kissinger2012}. Furthermore, radar sensors often do not fully observe an object at all viewing angles. This results in \emph{partial observability} in the radar response that introduces uncertainties in the reconstruction process. In this work, we tackle this difficult problem of object reconstruction from long-range, high-frequency radar responses that are partially observed. 

Previous deep learning approaches to long-range high-frequency radar modeling have focused primarily on extracting high-level features for classification, segmentation, or pose estimation tasks~\cite{muthukrishnan2023invrt,afit1999pose}, which are still difficult open problems. However, many downstream tasks require full-shape reconstruction of an observed object, a higher dimensional and even more challenging problem. The computer vision community has separately developed models for 3D shape reconstruction -- these models are typically conditioned on partial point clouds or multi-view images, and they often try to estimate camera intrinsic and extrinsic parameters. However, there are unique challenges when conditioning on a radar signal for 3D reconstruction instead of multi-view images. First, radar lacks analogous camera parameters to estimate because the observed shape does not correspond to simple geometric projections of the radar response. Second, the partial observability introduces high uncertainty. Much more of the object may be occluded from a radar's line of sight compared to a camera's single view. Additionally, the radar signal is spread across multiple frequencies, which can correspond to different resolutions of the object's geometry. Previous methods have not focused on the difficult problem of full-shape reconstruction from long-range high-frequency radar signals or taken advantage of individual frequencies in the radar response.

We propose \ours{}, a method that can reconstruct full 3D shapes from high-frequency radar responses by associating radar frequencies with shape resolutions. Our approach consists of two stages: 1) learning a multi-resolution, hierarchical latent space for 3D shapes, and 2) training a diffusion model to denoise in this space by conditioning on radar responses. The first stage uses a series of encodings and a VAE to learn a regularized latent space of vectors defining signed distance functions (SDFs)~\citep{chou2023diffusion}. Instead of representing a shape with a single latent vector, we separate the latent vector into components that represent shape features at multiple resolutions (e.g. the thin structures in a shape versus its overall structure). This representation is created by projecting multi-resolution point cloud features onto triplanes of various spatial resolutions. These are then processed separately and combined as input into an SDF network. The second stage uses a Transformer backbone to predict the denoised sequence of latent shape vectors, iteratively from coarsest to finest resolution, conditioned with attention on embeddings of the corresponding radar resolution. Additionally, we incorporate a domain-relevant 2D shape prior~\citep{muthukrishnan2023invrt} and propose a more efficient version of our method for this lower dimensional shape space by 1) encoding the shape space as a projection of our 2D shape parameterization and 2) using a U-Net to jointly encode the radar response and predict denoised latent shape vectors. Overall, we make four primary contributions:

\begin{itemize}
    \item We present \ours{}, a novel denoising diffusion model that reconstructs an object's 3D geometry from partially-observable, high-frequency radar observations. 
    \item We show superior results compared to many 3D reconstruction models adapted to the radar domain and an existing competitive radar baseline. 
    \item We demonstrate a general method for learning multiresolution signed distance functions of 3D geometries. 

    \item We introduce the \emph{Manifold40-PO} and \emph{Manifold40-PO-SBR} benchmark datasets, the first public datasets of diverse meshes and simulated high-frequency radar responses for radar-based single object reconstruction. 

\end{itemize}

\section{Related Work}

\subsection{Diffusion Models}
Diffusion models~\citep{ho2020denoising} have emerged as a powerful generative model applicable in many scientific domains ranging from bioinformatics~\citep{guo2024diffusion} to climate science~\citep{li2024generative, bassetti2024diffesm}. Alternative methods for generative modeling include Generative Adversarial Networks (GANs)~\cite{truong2019}, but diffusion has been shown to outperform GANs~\citep{rombach2022high, dhariwal2021diffusion}. Flow Matching~\citep{lipman2023flow} has also recently emerged as an alternative, but diffusion's demonstrated versatility across domains and its robustness to noisy inputs~\citep{webber2024diffusion} motivates its use in this work.

\subsection{3D Representations and Reconstruction} 
\label{subsec:3drepresentation}
3D reconstruction is a long-standing task in computer graphics and computer vision, leading to the development of many deep learning methods that take as input 2D shape projections (e.g., images, radar) and reconstruct the 3D geometries. In many use cases, entire scenes, consisting of geometries, lighting, transparency, density, and textures, must be modeled. Techniques like Gaussian Splatting~\citep{kerbl20233d} and Neural Radiance Fields~\citep{mildenhall2021nerf,klinghoffer2024platonerf} (NeRF) excel at modeling these high dimensional structures, but they are optimized for rendering, and extracting meshes from these representations is not straightforward. Instead, much research has focused on triplane features, point clouds, meshes, voxel, or signed distance function (SDF) representations for meshes~\cite{chan2022efficient,Girdhar16, Kar15, muthukrishnan2023invrt, Sharma16,Wu2016,arshad2023list,chou2023diffusion,park2019}.

Among these representations, Deep SDFs~\citep{park2019deepsdf} and subsequent improvements have gained popularity due to their efficiency and small memory footprint. Two-stage Diffusion~\citep{ho2020denoising} approaches have demonstrated success in generating these SDFs~\citep{chou2023diffusion,zheng2023locally,xiong2025octfusion,gao2025hieroctfusion}. Multiresolution hash encodings have improved performance of SDF-based representations by projecting the coordinates of query points to a higher dimensional, spatially-aware feature~\citep{muller2022instant}. These models are typically conditioned on 2D images or partial point clouds, and some use Octree-based structures to generate hierarchical features. However, none of these works are conditioned on radar observations, and the Octree's hierarchical features require manual part segmentation of the geometries as training labels. In this work, we use SDFs and learn hierarchical features without part segmentation labels by utilizing the multiresolution hash encoding in a novel way. 

\subsection{Radar Modeling with Deep Learning}
\label{paragraph:radar_related_work}

Many existing deep learning methods for radar modeling have used tools like NeRF and Gaussian splatting to extract geometric representations from autonomous driving data~\citep{borts2024radar,kung2025radarsplat}, but they require per-scene optimization during inference. There are existing methods that do not require per-scene optimization~\citep{zhang2024towards, mopidevi2024rmap, geng2024dream}, but these focus on reconstructing point-clouds, or use preprocessed radar data like NuScenes~\citep{caesar2020nuscenes} point clouds as input. None of these works solve the difficult problem of full mesh reconstruction from unprocessed radar data, and they also do not focus on long-range radar signals, where signal interference between closely spaced objects becomes minimal and single-object reconstruction is feasible. Instead, this work focuses on \emph{full mesh reconstruction} from \emph{raw, long-range, high-frequency} radar signals without per-scene optimization.

Within this domain of high-frequency radar, many existing deep learning algorithms infer object class rather than full shape. They typically encode spatial information using 1D convolutional neural networks (CNNs) \cite{hrrp_cnn, 
% hrrp_cnn2, hrrp_cnn3, hrrp_atr1, hrrp_cnn_attention, 
hrrp_cnn_attention2} or recurrent neural networks (RNNs) \cite{hrrp_rnn_attention}. Some of these approaches also apply attention to spatial encodings \cite{hrrp_cnn_attention, hrrp_cnn_attention2, hrrp_rnn_attention}, increasing model performance. InvRT~\cite{muthukrishnan2023invrt}, a custom transformer model, was designed to encode both the spatial and temporal structure of the radar signature to reconstruct the shape of roll-symmetric objects. The few methods that focus on full single-shape reconstruction do so for human body meshes~\citep{xue2021mmmesh, chen2022mmbody, zhao2019through}, but rely on parametric body priors, like SMPL~\citep{loper2023smpl}, to reconstruct meshes. 

The radar-based reconstruction of 3D shapes with \emph{arbitrary} topology remains a challenging task with high sensitivity across geometries in noisy, partially observable settings. In addition, there is a lack of diverse high-frequency radar datasets that can be used to train robust deep learning models for radar-based reconstruction. This gap is likely because existing real high-frequency radar data is heavily restricted by security and IP concerns, and capturing such data at high fidelity requires access to specialized equipment. In order to drive future research and reduce the barrier to entry for single-shape reconstruction methods from high-frequency radar, we introduce two large-scale datasets of \emph{diverse} geometries and simulated radar responses. We also tackle full 3D shape reconstruction and evaluate \ours{} against noisy, partially observable, and real radar responses to observe robustness.

\begin{figure*}
    \centering
    \includegraphics[width=\linewidth]{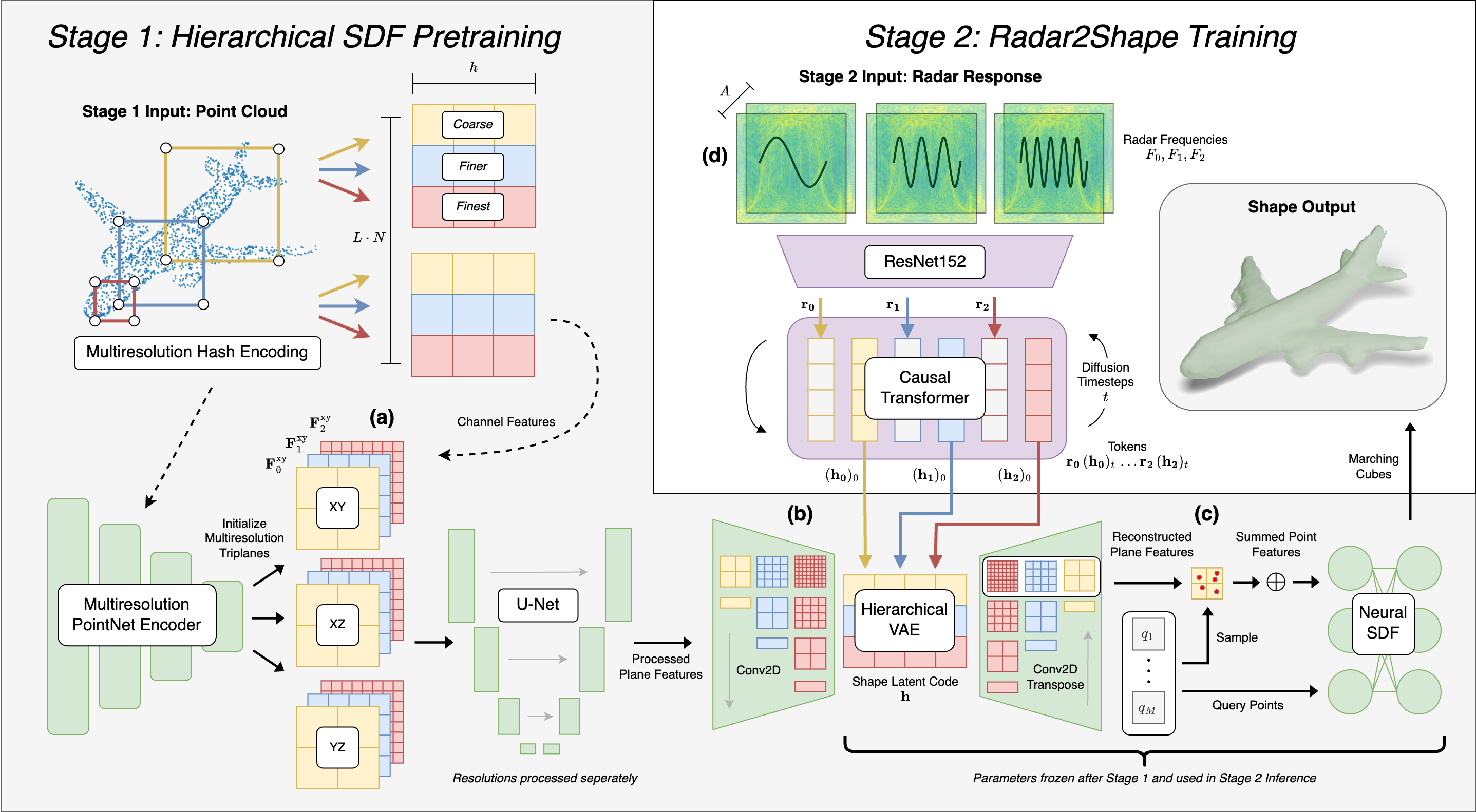}
    \caption{\textbf{Method.} \ours{} consists of two stages: 1) learning a multi-resolution, hierarchical latent space for 3D shapes, and 2) training a diffusion model to denoise in this space by conditioning on radar responses. In this figure, three hierarchical levels ($L=3$) are shown. \textbf{(a)} In Stage 1, we learn per-point multiresolution features from a point cloud that are projected onto triplanes of $L$ different grid resolutions. \textbf{(b)} A VAE then reconstructs each triplane independently to keep feature resolutions separate in its latent space. \textbf{(c)} Features are combined across resolutions to reconstruct the 3D geometry. \textbf{(d)} In Stage 2, a Transformer learns a sequence of $L$ multiresolution radar embeddings from a radar response interleaved with the VAE's multiresolution latent shape features. This enable coarse-to-fine prediction in a conditional diffusion process. Green and purple modules represent parameters trained during Stage 1 and Stage 2, respectively.}
    \label{fig:overview}
    \vspace{-3mm}
\end{figure*}

\section{Background}
 In this section, we discuss two core background concepts for \ours{}: Denoising Diffusion Probabilistic models and the first-principles physics model that generates high-frequency radar signatures from 3D object scattering. More details for each topic are provided in Appendix A and B. 

\subsection{Denoising Diffusion Probabilistic Models}
\label{subsec:ddpm}

Denoising Diffusion Probabilistic Models (DDPMs)~\citep{ho2020denoising} are generative models that leverage a forward diffusion process and a reverse denoising process to generate samples. The forward process adds Gaussian noise to a clean data sample \( x_0 \) over \( T \) timesteps, creating noisy samples \( x_t \). The reverse process aims to recover the clean data distribution by progressively denoising \( x_t \). The training objective of DDPMs is to minimize the variational lower bound of the negative log-likelihood of the generated data to match the true data distribution, over all timesteps \( 1 \) to \( T \).

\subsection{High-Frequency Radar Simulation for Single Object Reconstruction}
\label{subsec:radarbackground}

The techniques for modeling radar signatures of 3D objects depend on the relative size of the object $l$ and the wavelength of the radar $\lambda$. Most commercial and defense-related applications use high-frequency radar waveforms, where the object size is much larger than the radar wavelength, and where multiple closely spaced objects can be resolved. Because of this, signal interference between objects is minimal and can be ignored~\citep{ozturk2002implementation}. Therefore, in this domain, \emph{single} object reconstruction from radar is feasible. A single object's scattering response can often be reduced to a summation of discrete scattering centers by taking advantage of the Geometric Theory of Diffraction (GTD)~\cite{keller1962geometrical}.  This reduction allows the use of parametric, component-based, scattering models that reduce radar modeling to summing over responses of individual components. Examples of components are discrete points, spheres, rings, and triangles of a mesh, the latter being the focus of this paper.

\section{Method}

In this section, we present \ours{}, our method for generating 3D geometries from radar observations. Our approach relies on raw radar responses (Section~\ref{subsec:radarsignal}). \ours{} consists of two stages: first, we learn a latent space for 3D geometries using a point cloud to SDF model (Section~\ref{method:heirarchical_sdf}). Second, we train a diffusion model to denoise in this latent space by conditioning on radar responses, then produce an SDF by feeding the generated latent vector into our SDF decoder and running Marching Cubes~\citep{lorensen1998marching} to obtain the mesh (Section~\ref{method:radar}). More details on each topic are in Appendix C and D.

\subsection{Radar Signal Input}
\label{subsec:radarsignal}

Figure~\ref{fig:figure1} demonstrates our problem setting. The input to the learning task is a collection of radar responses at different viewing directions, and the output is the proposed observed geometry. 
Let $\bm{u}$ be viewing direction $\bm{u} = (\sin\alpha \cos\phi, \sin\alpha \sin\phi, \cos\alpha)$, for $\alpha \in [0, \pi]$ and $\phi \in [0, 2\pi]$, representing the aspect and roll angles, respectively. The corresponding radar response $F(\bm{u}, f)$ of an object is a sequence of real and imaginary scattering responses calculated using a linear set of frequencies $\{f_i\} \in [f_{\text{min}}, f_{\text{max}}]$, where the bandwidth of the signal is $B = f_\text{min} - f_{\text{max}}$. The input to \ours~is the amplitude measurement, calculated by taking the magnitude of the real and imaginary components $F$ and converted into decibel scale following $20 \times \log_{10}(|F|)$. This scale smooths out large fluctuations in signal strength and allows the input to represent a large range of values. 

For general 3D geometries, we discretize $\alpha$ and $\phi$ into $N_\alpha$ and $N_\phi$ bins, respectively, such that $F \in \mathbb{R}^{N_\alpha \times N_\phi \times |\{f_i\}|}$. We also incorporate a roll-symmetric shape prior when comparing to the baseline method, InvRT, which considered only roll-symmetric shapes of the \emph{Frusta} dataset~\cite{muthukrishnan2023invrt}. 
In this case, the radar response $F$ is identical across all roll angles $\phi$, and it is sufficient to index $F$ by the aspect angle $\alpha$ and the frequency $f$ alone, such that $R \in \mathbb{R}^{N_\alpha \times |\{f_i\}|}$.

\subsection{Stage 1: Hierarchical SDF Training}
\label{method:heirarchical_sdf}

In the first stage denoted by the gray box in Figure~\ref{fig:overview}, the SDF model is trained together with an encoder mapping object meshes $\mathcal{M}$ to hierarchical latent SDF codes $\mathbf{h} \in \mathcal{H}$, which represent the geometry of $\mathcal{M}$ at different resolutions. We draw inspiration from Diffusion SDF's~\citep{chou2023diffusion} architecture that regularizes $\mathcal{H}$ for easier downstream diffusion training, but we disentangle the hierarchies in the regularized latent space which has not been done in previous work. 

For a given number of resolution levels $L$, batch size $B$, points $\mathbf{p_i} \in \mathbf{P}$ on the surface of $\mathcal{M}$ where $1 \leq i \leq N$, we first embed each point into hierarchical feature space $\mathbf{f}_i^{\text{(in)}} = \text{MultiRes}(\mathbf{p_i}), \ \mathbf{f}_i^{\text{(in)}} \in \mathbb{R}^{L \times h}$ using the multiresolution hash encoding, and linearly project to the model dimension $H$ such that $\mathbf{F}^{\text{(in)}} \in  \mathbb{R}^{B \times N \times L \times H}$. Each level's features are independently processed with a series of ResNet blocks with local pooling, resulting in $\mathbf{F}^{(\text{out})} \in  \mathbb{R}^{B \times N \times L \times H}$ that contains per-point spatial context. 

We then train a \emph{Hierarchical VAE} (Figure~\ref{fig:overview}.b.) to reconstruct triplanes generated from these features, while maintaining hierarchical levels. Each plane ($XY$, $XZ$, $YZ$) is initialized as a grid for each level $l \in L$ with spatial resolutions $R$ increasing by powers of $2$, from coarse to fine -- for \ours, we use $R \in \{8, 16, 32, 64 \}$. Each grid cell's value is the mean of features from points projecting orthographically into it:

\begin{equation}
\left(\mathbf{F}^{\pi}_{l}\right)_{u,v} = \frac{1}{|S_{u,v}^{\pi}|} \sum_{i \in S_{u,v}^{\pi}} (\mathbf{F}^{\text{(out)}}_l)_i
\end{equation}

This creates a sparse feature grid for each level $\mathbf{F}^{\pi}_{l} \in \mathbb{R}^{B \times C \times R \times R}$ where the channel dimension of each level's grid uniquely corresponds to the features at that resolution from the multiresolution hash encoding (Figure~\ref{fig:overview}.a.). A 2D U-Net then densifies each sparse grid independently, resulting in multiresolution triplane features $\{\mathbf{F}^{\text{xy}}_{l}, \mathbf{F}^{\text{yz}}_{l}, \mathbf{F}^{\text{xz}}_{l}\}$. The VAE independently reconstructs each resolution triplane feature  (Figure~\ref{fig:overview}.b.), maintaining the level dimension in its stochastic latent variables $\mu_{\mathbf{h}}, \sigma^2_{\mathbf{h}} \in \mathbb{R}^{L \times Z}$ where $Z$ is the latent shape dimension. We treat the all levels jointly as a distribution and apply the following KL-divergence loss: $\text{KL}(\mathcal{N}(\mu_{\mathbf{h}}, \sigma^2_{\mathbf{h}}) \ || \ \mathcal{N}(0, 0.25))$, where $\mathbf{h} \sim \mathcal{N}(\mu_{\mathbf{h}}, \sigma^2_{\mathbf{h}})$ with the reparameterization trick, $\mathbf{h} \in \mathbb{R}^{L \times Z}$, and $\mathbf{h}_l \in \mathbb{R}^{Z}$.

Query points $q_i \dots q_M$ are then used to grid sample each triplane for all resolutions $L$, then summed across planes and resolutions to create a rich per-point multiresolution shape representation $\pi_i$. Each point's $q_i$ coordinate position is concatenated with $\pi_i$ which is input into a SDF MLP (Figure~\ref{fig:overview}.c). The final objective becomes the sum of the L1 SDF prediction error and the KL-divergence loss, with no need for a traditional VAE reconstruction loss.

\begin{table*}[htbp]
    \tiny
    \caption{\textbf{Quantitative Results.} Test performance of models after training on the Manifold40-PO dataset, with partial observability models using training-time mask augmentations. Metrics are evaluated across 20 random heldout meshes in the zero noise setting. Test-time partial observability is applied using a randomly sampled mask on 70\% of the signal, with monoconic having a fixed masked for consistent evaluation. \ours{} largely outperforms TMNET and LIST which struggle to learn and also Diffusion-SDF which is competitive in this domain. \ours{} has relatively strong zero-shot generalization to Manifold40-PO-SBR and the real monoconic radar response.}
    \centering
    \vspace{-6pt}
    \begin{tabular}{llcccccc}
        \toprule
        & & \multicolumn{3}{c}{\textbf{Full Observability}} & \multicolumn{3}{c}{\textbf{Partial Observability}}  \\
        \cmidrule(lr){3-5} \cmidrule(lr){6-8} 
        
        Dataset & Model & CD ($\downarrow$) & IoU ($\uparrow$) & F-Score ($\uparrow$) & CD ($\downarrow$) & IoU ($\uparrow$) & F-Score ($\uparrow$)  \\
        \midrule
        \midrule

        \multirow{4}{*}{\textbf{Manifold40-PO}} & \ours & $\mathbf{64.47 \pm 86.87}$ & $\mathbf{0.51 \pm 0.19}$ & $\mathbf{0.22 \pm 0.11}$ & $\mathbf{44.72 \pm 58.54}$ & $\mathbf{0.59 \pm 0.27}$ & $\mathbf{0.27 \pm 0.18}$ \\
                                 & Diffusion-SDF & $508.92 \pm 386.10$ & $0.13 \pm 0.11$ & $0.05 \pm 0.03$ & $566.73 \pm 321.76$ & $0.10 \pm 0.06$ & $0.04 \pm 0.02$ \\
                                 & TMNet & $3501.14 \pm 391.20$  & $0.01 \pm 0.00$ & $0.02 \pm 0.02$ &  $9801.15 \pm 3164.11$ & $0.00 \pm 0.00$ & $0.01 \pm 0.01$ \\
                                 & LIST & $73599.60 \pm 10230.19$ & $0.00 \pm 0.00$ & $0.00 \pm 0.00$ & $79936.75 \pm 12058.06$ & $0.00 \pm 0.00$ & $0.00 \pm 0.00$ \\
                                 
        \midrule
        \midrule
        \multirow{4}{*}{\textbf{Manifold40-PO-SBR}} & \ours & $\mathbf{96.91 \pm 85.96}$ & $\mathbf{0.44 \pm 0.26}$ & $\mathbf{0.10 \pm 0.05}$  & $\mathbf{121.14 \pm 91.18}$ & $\mathbf{0.44 \pm 0.25}$ & $\mathbf{0.10 \pm 0.08}$ \\
                                 & Diffusion-SDF & $531.71 \pm 322.96$ & $0.12 \pm 0.10$ & $0.04 \pm 0.01$ & $456.92 \pm 284.12$ & $0.11 \pm 0.10$ & $0.05 \pm 0.02$ \\
                                 & TMNet    & $1235.35 \pm 773.37$ & $0.02 \pm 0.02$ & $0.02 \pm 0.04$ &  $1060.80 \pm 703.15$ & $0.02 \pm 0.02$ & $0.03 \pm 0.02$   \\
                                 & LIST  & $20723.39 \pm 2073.50$ & $0.00 \pm 0.00$ & $0.00 \pm 0.00$ & $20728.32 \pm 2080.41$ & $0.00 \pm 0.00$ & $0.00 \pm 0.00$ \\
        \midrule
        \midrule
        \multirow{4}{*}{\textbf{Monoconic}} & \ours & $\mathbf{8.403}$ & $\mathbf{0.811}$ & $\mathbf{0.139}$ & $\mathbf{40.647}$ & $\mathbf{0.681}$ & $\mathbf{0.104}$ \\
                                 & Diffusion-SDF & $40.510$ & 0.$712$  & $0.077$ & $504.227$ & $0.191$  & $0.047$ \\
                                 & TMNet  &   $229.233$  & $0.041$ &  $0.028$ & $916.115$  & $0.014$ &  $0.012$   \\
                                 & LIST & $564.531$ & $0.025$ & $0.007$ & $1189.351$ & $0.021$ & $0.016$ \\
        \bottomrule
    \end{tabular}
    \label{table:radar2shape_sdf_results}
    
\end{table*}

\begin{table*}[htbp]
    % \fontsize{8}{9}\selectfont
    \tiny
    \caption{\textbf{Quantitative Results for Roll-symmetric Shapes.} Test performance of \ours{} and InvRT after training on the Frusta dataset, with both models using training-time mask and noise augmentations for their respective observability and noise level. Metrics are evaluated across 20 random heldout meshes under different observability and noise conditions (low =$-80$dB, medium =$-60$dB, and high =$-40$dB), with test-time partial observability applied as the same randomly sampled masks for up to 70\% of the aspect. \ours{} outperforms InvRT across most metrics, notably with a larger performance gap in the difficult high-noise setting.}
    
\centering 
\vspace{-6pt}
    \begin{tabular}{llccc ccc}
        \toprule
        & & \multicolumn{3}{c}{\textbf{Full Observability}} & \multicolumn{3}{c}{\textbf{Partial Observability}} \\
        \cmidrule(lr){3-5} \cmidrule(lr){6-8}
        Noise & Model & IoU-R $(\uparrow)$ & IoU-S ($\uparrow$) & MATCH-S ($\downarrow$) & IoU-R ($\uparrow$) & IoU-S ($\uparrow$) & MATCH-S ($\downarrow$)\\
        \midrule
        \midrule
        
        \multirow{2}{*}{\textbf{Low}} & \ours  & $0.67 \pm 0.22$ & \bm{$0.73 \pm 0.24$} & \bm{$0.11 \pm 0.09$} & \bm{$0.62 \pm 0.24$} & \bm{$0.67 \pm 0.25$} & \bm{$0.12 \pm 0.12$} \\
                             & InvRT & \bm{$0.70 \pm 0.13$} & $0.66 \pm 0.20$ & $0.16 \pm 0.11$   & $0.61 \pm 0.25$ & $0.66 \pm 0.20$ & $0.18 \pm 0.21$ \\

        \midrule
        \midrule 
        \multirow{2}{*}{\textbf{Medium}}  & \ours  & \bm{$0.71 \pm 0.18$} & \bm{$0.76 \pm 0.20$} & \bm{$0.11 \pm 0.11$} & \bm{$0.66 \pm 0.21$} & \bm{$0.71 \pm 0.24$} & \bm{$0.12 \pm 0.11$} \\
                                & InvRT & $0.70 \pm 0.24$ & $0.64 \pm 0.18$ & $0.18 \pm 0.09$ & $0.63 \pm 0.23$ & $0.66 \pm 0.15$ & $0.19 \pm 0.12$ \\

        \midrule
        \midrule
        \multirow{2}{*}{\textbf{High}} & \ours  & \bm{$0.77 \pm 0.16$} & \bm{$0.79 \pm 0.17$} & \bm{$0.10 \pm 0.10$} & \bm{$0.70 \pm 0.19$} & \bm{$0.74 \pm 0.21$} & $\bm{0.14 \pm 0.12} $ \\
                              & InvRT & $0.70 \pm 0.20$ & $0.72 \pm 0.13$ & $0.26 \pm 0.22$     & $0.63 \pm 0.23$ & $0.67 \pm 0.17$ & $0.27 \pm 0.20$\\
                                
        \bottomrule
    \end{tabular}
    \label{table:rollsymmetric-results}
\end{table*}

\subsection{Stage 2: Radar-Conditional Generation}
\label{method:radar}

In the second stage, we jointly train a radar encoder $\Phi : F \mapsto \mathbf{r}$ and a denoising network $\Theta : \left( \mathbf{h}_l \right)_t , \mathbf{r}, t \mapsto \epsilon_t $ to predict denoised latent shape codes $\left( \mathbf{h}_l \right)_0$ in a coarse-to-fine manner along hierarchical levels. First, the radar response $F$ defined in Section~\ref{subsec:radarsignal} is split into $L$ linearly spaced blocks along the frequency dimension such that $F_j = \mathbb{R}^{N_{\alpha} \times N_{\phi} \times A}$ where $A=\frac{|\{f_i\}|}{L}$. We treat bin $j=0 \dots L$ as positions and add a sine-cosine positional encoding so $\Phi$ can distinguish inputs of different radar frequencies. We use a ResNet152~\cite{he2016deep} for $\Phi$ due to its ability to efficiently extract information from signals with two spatial dimensions, which are aspect and roll in $F$. Each block is encoded as $\mathbf{r}_j = \Phi ( F_j^{\text{(in)}} )$. To encourage robustness in partially-observable scenarios, we randomly mask between 0\% and 70\% of the aspect and roll dimensions during training such that the unmasked regions remain continuous. We use a Transformer~\citep{vaswani2017attention} to learn a sequence of interleaved low-to-high-frequency radar encodings and low-to-high resolution shape encodings (defined in Section~\ref{method:heirarchical_sdf}) as $\mathbf{r}_0,\mathbf{h}_0, \dots, \mathbf{r}_j, \mathbf{h}_l , \dots , \mathbf{r}_L,\mathbf{h}_L$, and apply a lower-triangular causal attention mask. Therefore, $\Phi$ predicts the noise of a shape feature at resolution $l$ \emph{only} by attending to shape features at a coarser resolution and frequencies of the radar response $j \leq l$, enabling coarse-to-fine prediction. 

The loss function is the mean squared error between the predicted noise and scheduled noise added to the odd tokens of the sequence. The denoising process is defined as:
\begin{equation}
\label{eq:denoisingprocess}
\left( \mathbf{h} \right)_{t-1} = \frac{1}{\sqrt{\alpha_t}} \left( \left( \mathbf{h} \right)_t - \frac{\sqrt{1 - \alpha_t}}{\sqrt{1 - \bar{\alpha}_t}}\, \Theta (\left( \mathbf{h} \right)_t, \mathbf{r}, t) \right)
\end{equation}

\paragraph{Incorporating Roll-Symmetric Shape Priors.}
\label{subsubsec:frustamethod}

For roll symmetric objects, we define a lower dimensional space of potential shapes. This space can be used to define a reduced parameter model and to output the radial profile, a sequence of coordinates $(\mathbf{r}, \mathbf{z})$ defining the outer bounds of a half cross-section. We encode this shape parameterization $\mathbf{h}_t$ with a single linear layer and ReLU to match the $\alpha$ dimension of $R$, then concatenate $\mathbf{h}_t$ with $R$ along the frequency dimension. A 1D U-Net~\citep{ronneberger2015u} then jointly encodes the noisy latent shape vector at timestep $t$ with the radar response during downsampling, and learns the noise prediction for $\mathbf{h}_t$ during upsampling.

\begin{figure}[htbp]
    \centering
    \includegraphics[width=\linewidth]{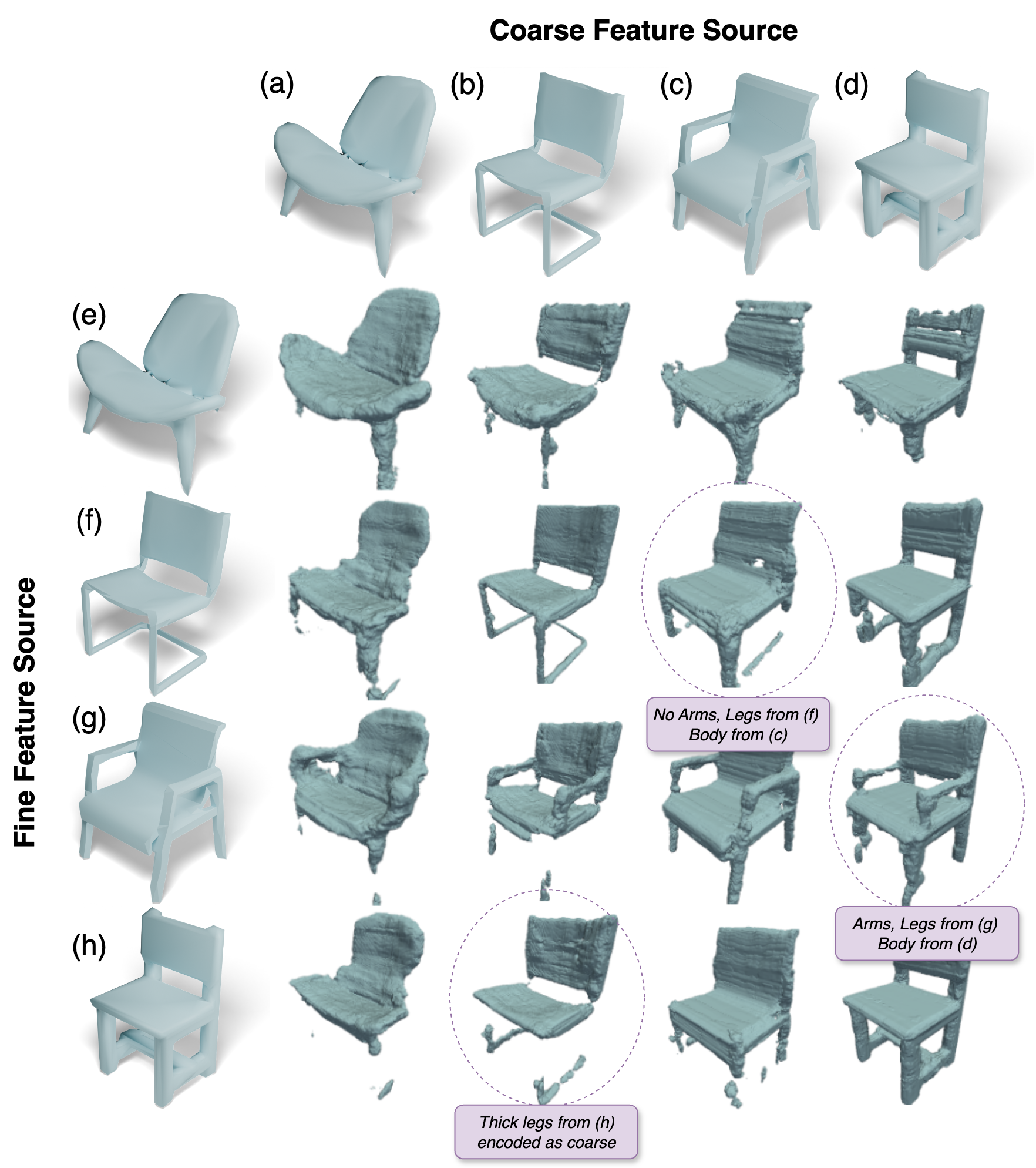}
    \caption{\textbf{Ablation.} Reconstruction of learned hierarchical latent codes with mixed coarse and fine features. For chairs, the model learns that the fine features correspond to arms and legs, because coarse features maintain the overall shape while the arms and legs are added or removed. This interpretability experiment demonstrates that our hierarchical SDF training method does indeed capture these coarse and fine features geometrically.}
    \label{fig:fig2_interp}
    \vspace{-4mm}
\end{figure}

\section{Experiments}
In this section, we describe baselines, metrics, and the data generation process, then present results of ~\ours{} on three benchmark datasets and real radar data. 

\paragraph{Baselines.}

We use three competitive image-to-shape models with tuned hyperparameters that span a variety of reconstruction methods to benchmark the performance of \ours{} on 3D reconstruction. TMNet~\citep{pan2019deep} iteratively transforms a topology to fit a target shape, and LIST~\citep{arshad2023list} uses spatial transformers for a coarse and fine prediction. Diffusion-SDF~\citep{chou2023diffusion} uses diffusion to predict SDFs, conceptually similar to our method. We adapt baselines to the radar problem by swapping their existing point cloud or image encoders for the same encoding network as \ours, a ResNet152, allowing equal comparison. We also choose these models for a fair comparison because they do not require camera estimation or pixel alignment, which would not make sense in the radar domain. For roll-symmetric geometries, we compare directly against the InvRT method, a transformer-based model that provides state-of-the-art performance on the roll-symmetric Frusta dataset~\citep{muthukrishnan2023invrt, kohler2023symmetric}.

\paragraph{Metrics of Evaluation.}

We measure the ability of~\ours{} to reconstruct general geometries by considering popular metrics for 3D mesh reconstruction~\citep{arshad2023list, pan2019deep, chen2025meshanything, li2021d2im, Choy16}: Chamfer distance (CD), intersection over union (IoU), and F-Score (1\%). For roll-symmetric shapes, the simplified $(\mathbf{r,z})$ parametrization enables accuracy to be evaluated more extensively than the metrics used for general geometries: \emph{IoU-S} measures the quality of shape predictions using the 2-dimensional binary mask intersection-over-union (IoU). (2) \emph{IoU-R} evaluates the ability of the predicted shapes to generate the ground truth radar phenomenology, and \emph{MATCH-S} evaluates the accuracy between ground truth and predicted shape segments by matching pairs of $(\mathbf{r,z})$. For further details on these metrics and their calculations, refer to Appendix E.

\subsection{Dataset Generation}

As discussed in Section~\ref{paragraph:radar_related_work}, there is a lack of diverse large-scale high-frequency radar datasets that can be used to train robust deep learning models for radar-based single-object reconstruction. We use the previously studied Frusta dataset to train and evaluate \ours{} against a competitive radar baseline, and although these shapes are domain-relevant, they are limited to 2D, roll-symmetric geometries. Therefore, we introduce \emph{Manifold40-PO}, the first publicly available, large-scale, high-frequency radar dataset which is generated from ModelNet40's~\citep{wu20153d} diverse set of over ten thousand unique real-world meshes. We rely on a widely accepted first-principles simulator using Physical Optics (PO)~\citep{balanis2012advanced} to generate radar responses, and use the \emph{Manifold40}~\citep{hu2022subdivision} variant of ModelNet40 for it's advantageous simulation properties. 

We also introduce a benchmark evaluation dateset with higher-fidelity effects, like multi-bounce interactions, using the Physical Optics and Shooting and Bouncing Rays (PO-SBR) algorithm~\citep{18706}, which we refer to as \emph{Manifold40-PO-SBR}. This algorithm is computationally expensive, so we only generate approximately two thousand samples for fine-tuning and evaluation to show that \ours{} can generalize when trained with Manifold40-PO. 

Real data in this radar domain, and equipment to record such data, is heavily restricted by cost, security and IP concerns. To our knowledge, there is no publicly available real high-frequency radar responses of single meshes. However, for this work, we are able to obtain real radar measurements across varying viewing angles of a monoconic object introduced in \cite{937474}. We evaluate \ours{} on these measurements after training on Manifold40-PO, and we make this data publicly available for future benchmarking. For further details on the monoconic object and simulation, see Appendix F and G.

\subsection{Performance on SDF Reconstruction}

\vspace{3mm}

\paragraph{Validating learned shape representations.}

To validate that Stage 1 learns to represent shape latent vectors at different shape resolutions, Figure~\ref{fig:fig2_interp} shows the meaning of these latent resolution levels from a geometric view. For chairs, the model learns that the fine features correspond to arms and legs, because the figure shows that coarse features maintain the overall shape while the arms and legs are added or removed. Note that in the bottom row, the legs of (h) are not substituted into the other shapes. This does not indicate failure, but is likely a mixing of geometric features across multiple granularity dimensions in the latent code, so unlike other samples, the legs of (h) may be encoded along with the shape's coarse feature. This is correct, and it can be expected because the legs appear thicker than other chairs, so they can be represented more coarsely. Stage 2 can then ``piece together" these shapes part-by-part in a coarse-to-fine manner, without any ground truth segmentation labels.

\paragraph{Comparison to baselines.}

\begin{figure}
    \centering
    \includegraphics[width=\linewidth]{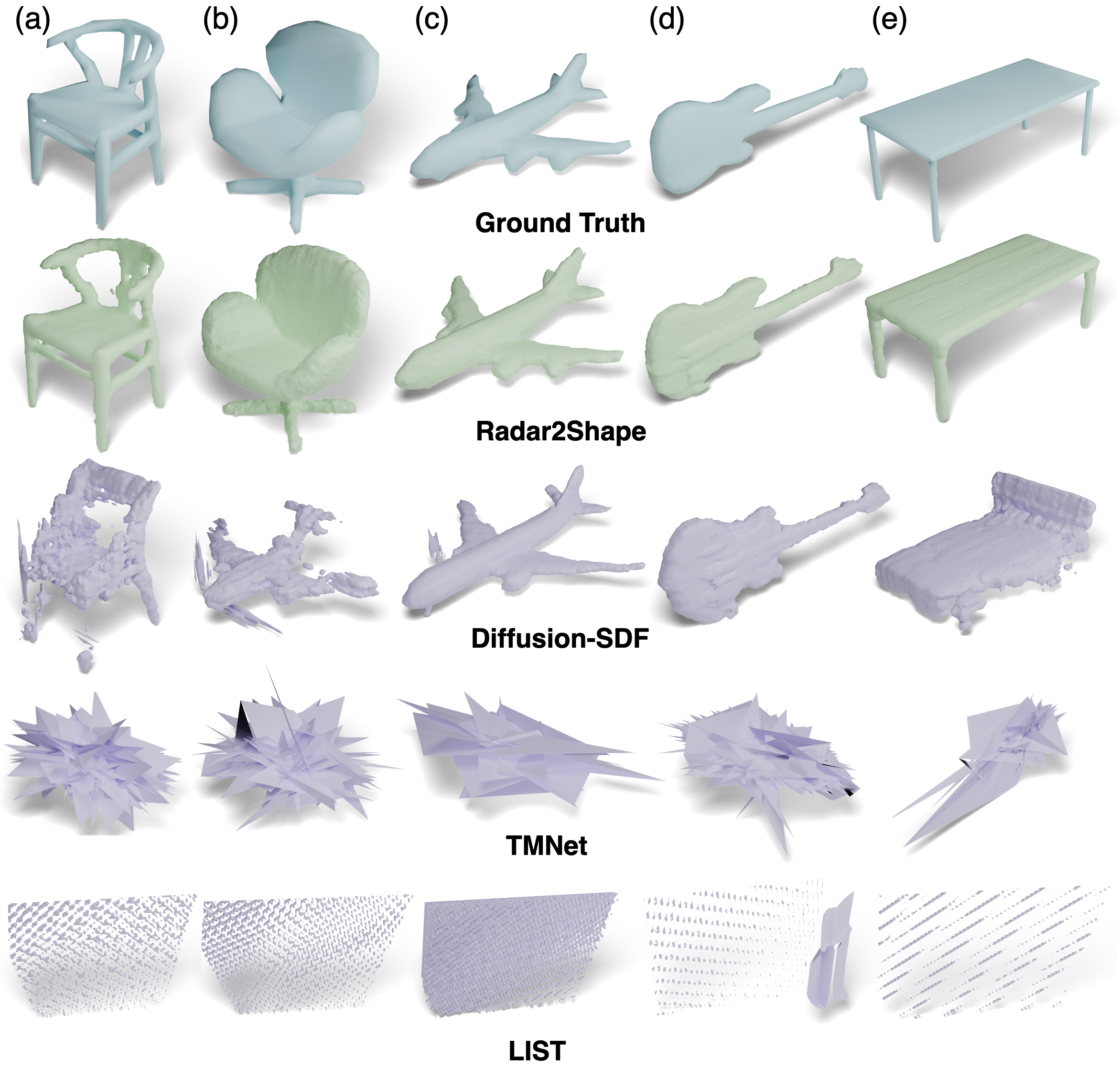}
    \caption{\textbf{Qualitative Results.} Comparison of select reconstructions from heldout fully-observed radar responses of Manifold40-PO. \ours{} consistently outperforms all baselines across a diverse set of meshes. TMNet and LIST exhibit mode-collapse, showing the difficulty of the radar-based 3D reconstruction problem when adapted to deterministic single/multi-view image-based reconstruction methods. Diffusion-SDF does the best among baselines, but often fails at reconstructing low-level features (shown with the chairs, table legs, and number of airplane engines).}
    \vspace{-5mm}
    \label{fig:reconstructions}
\end{figure}

\ours{}'s two-stage training approach greatly outperforms two competitive multi-view reconstruction methods, as shown in Figure~\ref{fig:reconstructions}. TMNet is unable to learn sphere deformations of sharp local features from the radar response, although it can generally learn the relative dimensions of the shape. LIST typically correlates query points with local image features, but since radar responses have different geometric dimensions than images, the model is unable to learn. This performance demonstrates the necessity for \ours{}, which geometrically leverages the frequencies of the radar response. 

We further observe the benefit of \ours{}'s coarse-to-fine refinement technique geometrically by comparing against the strongest baseline, Diffusion-SDF. Figure~\ref{fig:reconstructions}.b shows how Diffusion-SDF incorrectly reconstructs an airplane shape from the radar response of a chair. However, Diffusion-SDF still extracts the overall upward bending shape of the chair's arms and backrest, which it decodes as upward bending wings and a tail fin. Instead, \ours{} can first classify the shape correctly as a chair using the lowest frequency of the radar response and coarsest SDF resolution, then can focus on fine-grained features like the arms and backrest for an accurate reconstruction. 

Figure~\ref{fig:figure1} demonstrates the largest advantage of our method -- in partial observability. Stage 1 of Diffusion-SDF learns a latent space which struggles to disentangle lower level shape features from the overarching structure, so the latent space for diffusion tends to represent entire objects. If Stage 2's encoder learns a feature that represents only part of an object, as is often the case in partially-observable radar responses, it may not be representable in the latent space but might be ``nearest" to an entire object with similar features -- this results in high-variance guesses (Figure~\ref{fig:figure1}.c.). Instead, \ours{} first learns a disentangled latent space with hierarchical features (Figure~\ref{fig:fig2_interp}), allowing the radar features to be learned in accordance with the observed signal (Figure~\ref{fig:figure1}.b.). Table~\ref{table:radar2shape_sdf_results} reports reconstruction accuracy, where \ours{} also largely outperforms baselines across aggregate reconstruction metrics. 

\paragraph{Fine Tuning on Higher Fidelity Simulation.}

We show the zero-shot generalization results to Manifold40-PO-SBR in Table~\ref{table:radar2shape_sdf_results}, but to extract additional performance, we also fine-tune on a portion of the generated data while maintaining heldout samples for evaluation. We use LoRA~\citep{hu2022lora} fine-tuning on query and value attention projections, instead of full fine-tuning, to demonstrate that domain adaptation requires only lightweight changes to our model pretrained on the Manifold40-PO dataset. We observe a slight improvement in the fully observed setting, but modest improvement of about $0.07$ IoU and F-Score in the partial setting, suggesting high-fidelity artifacts may be more important when shape information is sparse. Further discussion and quantitive metrics are in Appendix H.

\subsection{Performance on Roll-symmetric Shapes}

Table~\ref{table:rollsymmetric-results} compares the U-Net variation of \ours{} (Section~\ref{subsubsec:frustamethod}) against InvRT across a variety of signal noise and observability settings. MATCH-S scores degrade for InvRT as noise increases, while \ours{} maintains performance. Since more noise creates an increasingly ill-posed problem, this performance gap demonstrates the advantage of diffusion as an inherently probabilistic model compared to a Transformer. Additionally, IOU-S provides an analogous metric to 3D IoU in Table~\ref{table:radar2shape_sdf_results}, with a modest performance gap between Manifold40-PO and Frusta performance in partial and full observability. This demonstrates that incorporating a roll-symmetric shape prior indeed improves shape reconstructions. Appendix I contains analysis on common failure cases and distributional accuracy.

\subsection{Application to Real Radar Data}
\label{subsec:realexperiment}

\begin{figure}[t]
    \centering
    \includegraphics[height=20mm,width=0.7\linewidth]{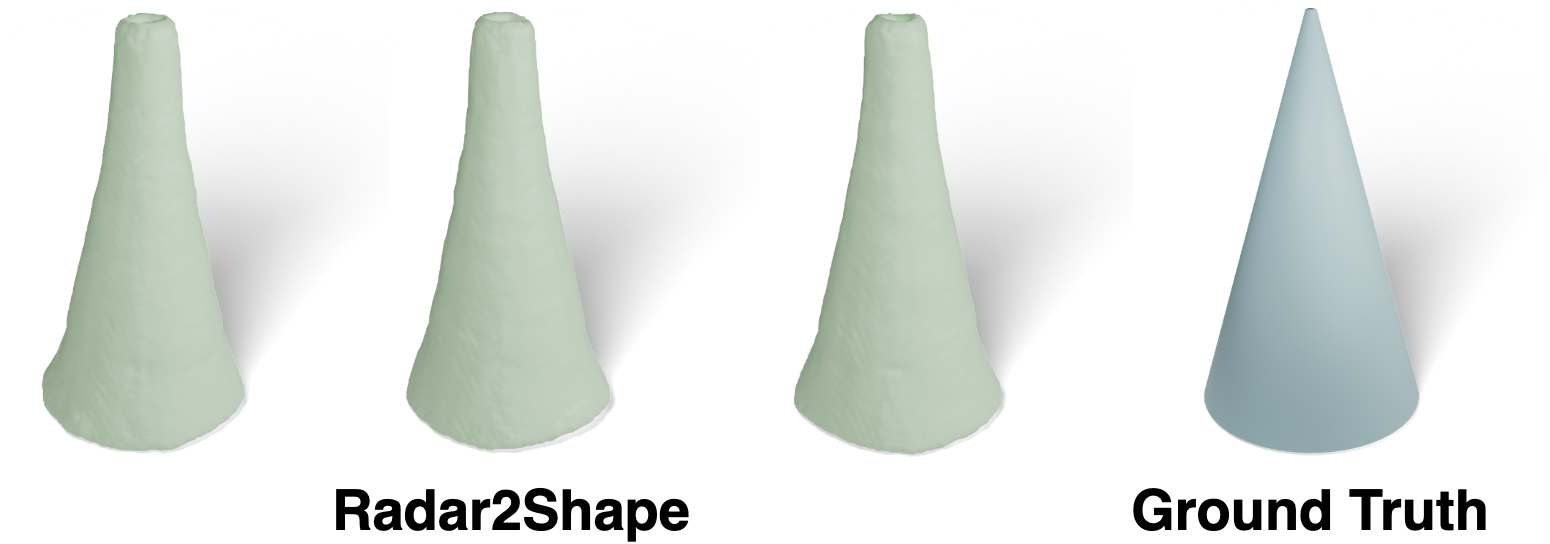}
    \caption{\textbf{Qualitative Results on Real Data.} Reconstructions of a monoconic object from its \emph{real} radar response, using \ours{} trained on Manifold40-PO. \ours{} predicts a wider tip, but is able to correctly predict the overall shape, base width, height, and angle near the base with low variance.}
    \vspace{-4mm}
    \label{fig:monoconic_reconstructions}
\end{figure}

To test zero-shot generalization properties of \ours, we consider real radar measurement data of a monoconic object introduced in~\cite{937474}. Since this object is roll-symmetric, its measurements are taken only along the aspect dimension $\alpha$. Although we could use the roll-symmetric variation of \ours{}, we choose to demonstrate the harder problem of full 3D reconstruction. Therefore, the input to \ours{} becomes the measurement repeated along roll angle $\phi$. Figure~\ref{fig:monoconic_reconstructions} and Table~\ref{table:radar2shape_sdf_results} (Monoconic) show the results of single-shot generalization to this object, given that \ours{} is trained on Manifold40-PO objects. \ours{} struggles with the tip of the cone likely due to the real data being recorded at a different object scale, but even with this slight distribution shift, it exhibits the best performance compared to the other baselines and reconstructs elements like base width accurately.

\section{Conclusion}

We present \ours{}, a novel method that can reconstruct 3D shapes from radar responses by associating signal frequencies with multiresolution shape features. We empirically demonstrate that this method proves to be more accurate than previous work, especially in noisy and partially observable settings. This work also introduces a general method to learn multiresolution signed distance functions, and establishes two benchmark datasets consisting of diverse meshes and high-frequency radar responses to drive future research in high-frequency radar modeling. 

\paragraph{Limitations and Future Work.}

There are some limitations to this work. We find that the hierarchical features learned in Stage 1 can be spatially bound (e.g. if pose changes, the fine-grained representation of a chair's leg might change its shape), which future work could mitigate by using rotation invariance. \ours{} does not attempt to learn the scale of reconstructed objects, since ModelNet40 does not represent relative scale among objects correctly. Future work can also collect more diverse real-world data to evaluate performance, fine-tuning if necessary as we have demonstrated with Manifold40-PO-SBR.

\section*{Acknowledgments}

This work is supported in part by NSF 2107256 and NSF 2134178. Tzofi Klinghoffer is supported by the Department of Defense (DoD) National Defense Science and Engineering Graduate (NDSEG) Fellowship Program. This material is also based on Neel Sortur's work supported by the NSF GRFP under Grant No. DGE-2439018.

{
    \small
    \bibliographystyle{ieeenat_fullname}
    \bibliography{main}

@article{geng2024dream,
  title={Dream-pcd: Deep reconstruction and enhancement of mmwave radar pointcloud},
  author={Geng, Ruixu and Li, Yadong and Zhang, Dongheng and Wu, Jincheng and Gao, Yating and Hu, Yang and Chen, Yan},
  journal={IEEE Transactions on Image Processing},
  year={2024},
  publisher={IEEE}
}

@inproceedings{mopidevi2024rmap,
  title={RMap: Millimeter-wave radar mapping through volumetric upsampling},
  author={Mopidevi, Ajay Narasimha and Harlow, Kyle and Heckman, Christoffer},
  booktitle={2024 IEEE/RSJ International Conference on Intelligent Robots and Systems (IROS)},
  pages={1108--1115},
  year={2024},
  organization={IEEE}
}

@article{zhang2024towards,
  title={Towards dense and accurate radar perception via efficient cross-modal diffusion model},
  author={Zhang, Ruibin and Xue, Donglai and Wang, Yuhan and Geng, Ruixu and Gao, Fei},
  journal={IEEE Robotics and Automation Letters},
  year={2024},
  publisher={IEEE}
}

@ARTICLE{18706,
  author={Ling, H. and Chou, R.-C. and Lee, S.-W.},
  journal={IEEE Transactions on Antennas and Propagation}, 
  title={Shooting and bouncing rays: calculating the RCS of an arbitrarily shaped cavity}, 
  year={1989},
  volume={37},
  number={2},
  pages={194-205},
  doi={10.1109/8.18706}}

@inproceedings{klinghoffer2024platonerf,
  title={PlatoNeRF: 3D reconstruction in Plato's cave via single-view two-bounce lidar},
  author={Klinghoffer, Tzofi and Xiang, Xiaoyu and Somasundaram, Siddharth and Fan, Yuchen and Richardt, Christian and Raskar, Ramesh and Ranjan, Rakesh},
  booktitle={Proceedings of the IEEE/CVF Conference on Computer Vision and Pattern Recognition},
  pages={14565--14574},
  year={2024}
}

@incollection{loper2023smpl,
  title={SMPL: A skinned multi-person linear model},
  author={Loper, Matthew and Mahmood, Naureen and Romero, Javier and Pons-Moll, Gerard and Black, Michael J},
  booktitle={Seminal Graphics Papers: Pushing the Boundaries, Volume 2},
  pages={851--866},
  year={2023}
}

@book{ozturk2002implementation,
  title={Implementation of physical theory of diffraction for radar cross section calculations},
  author={{\"O}zt{\"u}rk, Alper K{\"u}r{\c{s}}at},
  year={2002},
  publisher={Bilkent Universitesi (Turkey)}
}

@inproceedings{lipman2023flow,
  title={Flow Matching for Generative Modeling},
  author={Lipman, Yaron and Chen, Ricky T. Q. and Ben-Hamu, Heli and Nickel, Maximilian and Le, Matthew},
  booktitle={International Conference on Learning Representations (ICLR) 2023},
  year={2023},
  url={https://openreview.net/forum?id=PqvMRDCJT9t}
}

@inproceedings{borts2024radar,
  title={Radar fields: Frequency-space neural scene representations for fmcw radar},
  author={Borts, David and Liang, Erich and Broedermann, Tim and Ramazzina, Andrea and Walz, Stefanie and Palladin, Edoardo and Sun, Jipeng and Brueggemann, David and Sakaridis, Christos and Van Gool, Luc and others},
  booktitle={ACM SIGGRAPH 2024 Conference Papers},
  pages={1--10},
  year={2024}
}

@inproceedings{kung2025radarsplat,
  title={Radarsplat: Radar gaussian splatting for high-fidelity data synthesis and 3d reconstruction of autonomous driving scenes},
  author={Kung, Pou-Chun and Harisha, Skanda and Vasudevan, Ram and Eid, Aline and Skinner, Katherine A},
  booktitle={Proceedings of the IEEE/CVF International Conference on Computer Vision},
  pages={27596--27606},
  year={2025}
}

@inproceedings{
hu2022lora,
title={Lo{RA}: Low-Rank Adaptation of Large Language Models},
author={Edward J Hu and Yelong Shen and Phillip Wallis and Zeyuan Allen-Zhu and Yuanzhi Li and Shean Wang and Lu Wang and Weizhu Chen},
booktitle={International Conference on Learning Representations},
year={2022},
url={https://openreview.net/forum?id=nZeVKeeFYf9}
}

@inproceedings{zhao2019through,
  title={Through-wall human mesh recovery using radio signals},
  author={Zhao, Mingmin and Liu, Yingcheng and Raghu, Aniruddh and Li, Tianhong and Zhao, Hang and Torralba, Antonio and Katabi, Dina},
  booktitle={Proceedings of the IEEE/CVF International Conference on Computer Vision},
  pages={10113--10122},
  year={2019}
}

@inproceedings{xue2021mmmesh,
  title={mmMesh: Towards 3D real-time dynamic human mesh construction using millimeter-wave},
  author={Xue, Hongfei and Ju, Yan and Miao, Chenglin and Wang, Yijiang and Wang, Shiyang and Zhang, Aidong and Su, Lu},
  booktitle={Proceedings of the 19th Annual International Conference on Mobile Systems, Applications, and Services},
  pages={269--282},
  year={2021}
}

@inproceedings{chen2022mmbody,
  title={mmbody benchmark: 3d body reconstruction dataset and analysis for millimeter wave radar},
  author={Chen, Anjun and Wang, Xiangyu and Zhu, Shaohao and Li, Yanxu and Chen, Jiming and Ye, Qi},
  booktitle={Proceedings of the 30th ACM International Conference on Multimedia},
  pages={3501--3510},
  year={2022}
}

@inproceedings{caesar2020nuscenes,
  title={nuscenes: A multimodal dataset for autonomous driving},
  author={Caesar, Holger and Bankiti, Varun and Lang, Alex H and Vora, Sourabh and Liong, Venice Erin and Xu, Qiang and Krishnan, Anush and Pan, Yu and Baldan, Giancarlo and Beijbom, Oscar},
  booktitle={Proceedings of the IEEE/CVF conference on computer vision and pattern recognition},
  pages={11621--11631},
  year={2020}
}

@article{zheng2023locally,
  title={Locally attentional sdf diffusion for controllable 3d shape generation},
  author={Zheng, Xin-Yang and Pan, Hao and Wang, Peng-Shuai and Tong, Xin and Liu, Yang and Shum, Heung-Yeung},
  journal={ACM Transactions on Graphics (ToG)},
  volume={42},
  number={4},
  pages={1--13},
  year={2023},
  publisher={ACM New York, NY, USA}
}

@article{gao2025hieroctfusion,
  title={HierOctFusion: Multi-scale Octree-based 3D Shape Generation via Part-Whole-Hierarchy Message Passing},
  author={Gao, Xinjie and Du, Bi'an and Hu, Wei},
  journal={arXiv preprint arXiv:2508.11106},
  year={2025}
}

@inproceedings{xiong2025octfusion,
  title={OctFusion: Octree-based Diffusion Models for 3D Shape Generation},
  author={Xiong, Bojun and Wei, Si-Tong and Zheng, Xin-Yang and Cao, Yan-Pei and Lian, Zhouhui and Wang, Peng-Shuai},
  booktitle={Computer Graphics Forum},
  volume={44},
  number={5},
  pages={e70198},
  year={2025},
  organization={Wiley Online Library}
}

@article{muller2022instant,
  title={Instant neural graphics primitives with a multiresolution hash encoding},
  author={M{\"u}ller, Thomas and Evans, Alex and Schied, Christoph and Keller, Alexander},
  journal={ACM transactions on graphics (TOG)},
  volume={41},
  number={4},
  pages={1--15},
  year={2022},
  publisher={ACM New York, NY, USA}
}

@ARTICLE{937474,
  author={Mayhan, J.T. and Burrows, M.L. and Cuomo, K.M. and Piou, J.E.},
  journal={IEEE Transactions on Aerospace and Electronic Systems}, 
  title={High resolution 3D "snapshot" ISAR imaging and feature extraction}, 
  year={2001},
  volume={37},
  number={2},
  pages={630-642},
  doi={10.1109/7.937474}}

@inproceedings{wu20153d,
  title={3D ShapeNets: A Deep Representation for Volumetric Shapes},
  author={Wu, Zhirong and Song, Shuran and Khosla, Aditya and Yu, Fisher and Zhang, Linguang and Tang, Xiaoou and Xiao, Jianxiong},
  booktitle={Proceedings of the IEEE Conference on Computer Vision and Pattern Recognition (CVPR)},
  pages={1912--1920},
  year={2015},
  url={https://modelnet.cs.princeton.edu/}
}

@techreport{afit1999pose,
  title={Target Pose Estimation from Radar Data Using Adaptive Networks},
  author={Air Force Institute of Technology},
  year={1999},
  institution={Defense Technical Information Center},
  note={ADA361658}
}

@inproceedings{chan2022efficient,
  title={Efficient geometry-aware 3d generative adversarial networks},
  author={Chan, Eric R and Lin, Connor Z and Chan, Matthew A and Nagano, Koki and Pan, Boxiao and De Mello, Shalini and Gallo, Orazio and Guibas, Leonidas J and Tremblay, Jonathan and Khamis, Sameh and others},
  booktitle={Proceedings of the IEEE/CVF conference on computer vision and pattern recognition},
  pages={16123--16133},
  year={2022}
}

@article{mildenhall2021nerf,
  title={Nerf: Representing scenes as neural radiance fields for view synthesis},
  author={Mildenhall, Ben and Srinivasan, Pratul P and Tancik, Matthew and Barron, Jonathan T and Ramamoorthi, Ravi and Ng, Ren},
  journal={Communications of the ACM},
  volume={65},
  number={1},
  pages={99--106},
  year={2021},
  publisher={ACM New York, NY, USA}
}

@article{kerbl20233d,
  title={3d gaussian splatting for real-time radiance field rendering.},
  author={Kerbl, Bernhard and Kopanas, Georgios and Leimk{\"u}hler, Thomas and Drettakis, George},
  journal={ACM Trans. Graph.},
  volume={42},
  number={4},
  pages={139--1},
  year={2023}
}

@inproceedings{li2021d2im,
  title={D2im-net: Learning detail disentangled implicit fields from single images},
  author={Li, Manyi and Zhang, Hao},
  booktitle={Proceedings of the IEEE/CVF Conference on Computer Vision and Pattern Recognition},
  pages={10246--10255},
  year={2021}
}

@inproceedings{chen2025meshanything,
  title     = {MeshAnything: Artist-Created Mesh Generation with Autoregressive Transformers},
  author    = {Yiwen Chen and Tong He and Di Huang and Weicai Ye and Sijin Chen and Jiaxiang Tang and Zhongang Cai and Lei Yang and Gang Yu and Guosheng Lin and Chi Zhang},
  booktitle = {International Conference on Learning Representations (ICLR)},
  year      = {2025},
  note      = {Poster},
  url       = {https://openreview.net/forum?id=KGZAs8VcOM}
}

@misc{park2019,
  author = {Park, Jeong Joon and Florence, Peter and Straub, Julian and Newcombe, Richard and Lovegrove, Steven},
  description = {[1901.05103] DeepSDF: Learning Continuous Signed Distance Functions for Shape Representation},
  note = {cite arxiv:1901.05103},
  title = {DeepSDF: Learning Continuous Signed Distance Functions for Shape
  Representation},
  url = {http://arxiv.org/abs/1901.05103},
  year = 2019
}

@article{Sharma16,
  author       = {Abhishek Sharma and
                  Oliver Grau and
                  Mario Fritz},
  title        = {VConv-DAE: Deep Volumetric Shape Learning Without Object Labels},
  journal      = {CoRR},
  volume       = {abs/1604.03755},
  year         = {2016},
  url          = {http://arxiv.org/abs/1604.03755},
  eprinttype    = {arXiv},
  eprint       = {1604.03755},
  bibsource    = {dblp computer science bibliography, https://dblp.org}
}

@article{Girdhar16,
  author       = {Rohit Girdhar and
                  David F. Fouhey and
                  Mikel Rodriguez and
                  Abhinav Gupta},
  title        = {Learning a Predictable and Generative Vector Representation for Objects},
  journal      = {CoRR},
  volume       = {abs/1603.08637},
  year         = {2016},
  url          = {http://arxiv.org/abs/1603.08637},
  eprinttype    = {arXiv},
  eprint       = {1603.08637},
  bibsource    = {dblp computer science bibliography, https://dblp.org}
}

@article{Choy16,
  author       = {Christopher B. Choy and
                  Danfei Xu and
                  JunYoung Gwak and
                  Kevin Chen and
                  Silvio Savarese},
  title        = {3D-R2N2: {A} Unified Approach for Single and Multi-view 3D Object
                  Reconstruction},
  journal      = {CoRR},
  volume       = {abs/1604.00449},
  year         = {2016},
  url          = {http://arxiv.org/abs/1604.00449},
  eprinttype    = {arXiv},
  eprint       = {1604.00449},
  bibsource    = {dblp computer science bibliography, https://dblp.org}
}

@article{Wu2016,
  title={Single Image 3D Interpreter Network},
  author={Jiajun Wu and Tianfan Xue and Joseph J. Lim and Yuandong Tian and Joshua B. Tenenbaum and Antonio Torralba and William T. Freeman},
  journal={ArXiv},
  year={2016},
  volume={abs/1604.08685},
  url={https://api.semanticscholar.org/CorpusID:2003389}
}

@inproceedings{Kar15,
  author = {Kar, Abhishek and Tulsiani, Shubham and Carreira, João and Malik, Jitendra},
  booktitle = {CVPR},
  ee = {https://doi.ieeecomputersociety.org/10.1109/CVPR.2015.7298807},
  isbn = {978-1-4673-6964-0},
  pages = {1966-1974},
  publisher = {IEEE Computer Society},
  title = {Category-specific object reconstruction from a single image.},
  url = {http://dblp.uni-trier.de/db/conf/cvpr/cvpr2015.html#KarTCM15},
  year = 2015
}

@inproceedings{he2016deep,
  title={Deep Residual Learning for Image Recognition},
  author={He, Kaiming and Zhang, Xiangyu and Ren, Shaoqing and Sun, Jian},
  booktitle={Proceedings of the IEEE Conference on Computer Vision and Pattern Recognition (CVPR)},
  pages={770--778},
  year={2016}
}

@article{hu2022subdivision,
  title={Subdivision-based mesh convolution networks},
  author={Hu, Shi-Min and Liu, Zheng-Ning and Guo, Meng-Hao and Cai, Jun-Xiong and Huang, Jiahui and Mu, Tai-Jiang and Martin, Ralph R},
  journal={ACM Transactions on Graphics (TOG)},
  volume={41},
  number={3},
  pages={1--16},
  year={2022},
  publisher={ACM New York, NY}
}

@inproceedings{chou2023diffusion,
  title={Diffusion-sdf: Conditional generative modeling of signed distance functions},
  author={Chou, Gene and Bahat, Yuval and Heide, Felix},
  booktitle={Proceedings of the IEEE/CVF international conference on computer vision},
  pages={2262--2272},
  year={2023}
}

@inproceedings{pan2019deep,
  title={Deep mesh reconstruction from single rgb images via topology modification networks},
  author={Pan, Junyi and Han, Xiaoguang and Chen, Weikai and Tang, Jiapeng and Jia, Kui},
  booktitle={Proceedings of the IEEE/CVF International Conference on Computer Vision},
  pages={9964--9973},
  year={2019}
}

@inproceedings{arshad2023list,
  title={LIST: learning implicitly from spatial transformers for single-view 3D reconstruction},
  author={Arshad, Mohammad Samiul and Beksi, William J},
  booktitle={Proceedings of the IEEE/CVF International Conference on Computer Vision},
  pages={9321--9330},
  year={2023}
}

@article{ronneberger2015u, 

    title={U-Net: Convolutional Networks for Biomedical Image Segmentation}, 

    author={Ronneberger, Olaf and  Fischer, Philipp and Brox, Thomas}, 

    journal={Medical Image Computing and Computer-Assisted Intervention}, 

    year={2015}, 

    volume={19}, 

    number={3}, 

    pages={934-941}, 

    doi={10.1007/978-3-319-24810-3_154} 

}

@INPROCEEDINGS{truong2019,
  author={Truong, Thomas and Yanushkevich, Svetlana},
  booktitle={2019 International Joint Conference on Neural Networks (IJCNN)}, 
  title={Generative Adversarial Network for Radar Signal Synthesis}, 
  year={2019},
  volume={},
  number={},
  pages={1-7},
  doi={10.1109/IJCNN.2019.8851887}}

@article{keller1962geometrical,
  title={Geometrical theory of diffraction},
  author={Keller, Joseph B},
  journal={J. Opt. Soc. Am},
  volume={52},
  number={2},
  pages={1},
  year={1962}
}

@inproceedings{rombach2022high,
  title={High-resolution image synthesis with latent diffusion models},
  author={Rombach, Robin and Blattmann, Andreas and Lorenz, Dominik and Esser, Patrick and Ommer, Bj{\"o}rn},
  booktitle={Proceedings of the IEEE/CVF conference on computer vision and pattern recognition},
  pages={10684--10695},
  year={2022}
}

@article{dhariwal2021diffusion,
  title={Diffusion models beat gans on image synthesis},
  author={Dhariwal, Prafulla and Nichol, Alexander},
  journal={Advances in neural information processing systems},
  volume={34},
  pages={8780--8794},
  year={2021}
}

@article{bassetti2024diffesm,
  title={DiffESM: Conditional emulation of temperature and precipitation in Earth system models with 3D diffusion models},
  author={Bassetti, Seth and Hutchinson, Brian and Tebaldi, Claudia and Kravitz, Ben},
  journal={Journal of Advances in Modeling Earth Systems},
  volume={16},
  number={10},
  pages={e2023MS004194},
  year={2024},
  publisher={Wiley Online Library}
}

@article{li2024generative,
  title={Generative emulation of weather forecast ensembles with diffusion models},
  author={Li, Lizao and Carver, Robert and Lopez-Gomez, Ignacio and Sha, Fei and Anderson, John},
  journal={Science Advances},
  volume={10},
  number={13},
  pages={eadk4489},
  year={2024},
  publisher={American Association for the Advancement of Science}
}

@article{webber2024diffusion,
  title={Diffusion models for medical image reconstruction},
  author={Webber, George and Reader, Andrew J},
  journal={BJR| Artificial Intelligence},
  volume={1},
  number={1},
  pages={ubae013},
  year={2024},
  publisher={Oxford University Press}
}

@article{guo2024diffusion,
  title={Diffusion models in bioinformatics and computational biology},
  author={Guo, Zhiye and Liu, Jian and Wang, Yanli and Chen, Mengrui and Wang, Duolin and Xu, Dong and Cheng, Jianlin},
  journal={Nature reviews bioengineering},
  volume={2},
  number={2},
  pages={136--154},
  year={2024},
  publisher={Nature Publishing Group UK London}
}

@inproceedings{BergenBGD02a,
  author = {Bergen, Kathleen M. and Brown, Daniel G. and Gustafson, Eric J. and Dobson, M. Craig},
  booktitle = {DG.O},
  ee = {http://dl.acm.org/citation.cfm?id=1123167},
  publisher = {Digital Government Research Center},
  series = {ACM International Conference Proceeding Series},
  title = {Radar Remote Sensing of Habitat Structure for Biodiversity Informatics.},
  url = {http://dblp.uni-trier.de/db/conf/dgo/dgo2002.html#BergenBGD02a},
  year = 2002
}

@inproceedings{barnes2020,
author = {Barnes, Dan and Posner, Ingmar},
year = {2020},
month = {05},
pages = {9484-9490},
title = {Under the Radar: Learning to Predict Robust Keypoints for Odometry Estimation and Metric Localisation in Radar},
doi = {10.1109/ICRA40945.2020.9196835}
}

@Inbook{Kissinger2012,
author="Kissinger, D.",
title="Radar Fundamentals",
bookTitle="Millimeter-Wave Receiver Concepts for 77 GHz Automotive Radar in Silicon-Germanium Technology",
year="2012",
publisher="Springer US",
address="Boston, MA",
pages="9--19",
isbn="978-1-4614-2290-7",
doi="10.1007/978-1-4614-2290-7_2",
url="https://doi.org/10.1007/978-1-4614-2290-7_2"
}

@ARTICLE{hrrp_cnn_attention,
  author={Pan, M. and Liu, A. and Yu, Y. and Wang, P. and Li, J. and Liu, Y. and Lv, S. and Zhu, H.},
  journal={IEEE Trans. on Geoscience and Remote Sensing}, 
  title={Radar HRRP Target Recognition Model Based on a Stacked {CNN–Bi-RNN} With Attention Mechanism}, 
  year={2022},
  volume={60},
  number={},
  pages={1-14},
  doi={10.1109/TGRS.2021.3055061}}

@ARTICLE{hrrp_cnn_attention2,
  author={Wan, J. and Chen, B. and Liu, Y. and Yuan, Y. and Liu, H. and Jin, L.},
  journal={IEEE Access}, 
  title={Recognizing the {HRRP} by Combining {CNN} and {BiRNN} With Attention Mechanism}, 
  year={2020},
  volume={8},
  number={},
  pages={20828-20837},
  doi={10.1109/ACCESS.2020.2969450}}

@INPROCEEDINGS{hrrp_cnn,
  author={Lundén, J. and Koivunen, V.},
  booktitle={IEEE Radar Conf.}, 
  title={Deep learning for {HRRP}-based target recognition in multistatic radar systems}, 
  year={2016},
  volume={},
  number={},
  pages={1-6},
  doi={10.1109/RADAR.2016.7485271}}

@article{hrrp_rnn_attention,
title = {Target-Aware Recurrent Attentional Network for Radar {HRRP} Target Recognition},
journal = {Sig. Proc.},
volume = {155},
pages = {268-280},
year = {2019},
issn = {0165-1684},
doi = {https://doi.org/10.1016/j.sigpro.2018.09.041},
url = {https://www.sciencedirect.com/science/article/pii/S0165168418303220},
author = {B. Xu and B. Chen and J. Wan and H. Liu and L. Jin}
}

@ARTICLE{radar_autonomous_vehicles,
  author={Bilik, I. and Longman, O. and Villeval, S. and Tabrikian, J.},
  journal={IEEE Sig. Proc. Magazine}, 
  title={The Rise of Radar for Autonomous Vehicles: Signal Processing Solutions and Future Research Directions}, 
  year={2019},
  volume={36},
  number={5},
  pages={20-31},
  doi={10.1109/MSP.2019.2926573}}

@incollection{lorensen1998marching,
  title={Marching cubes: A high resolution 3D surface construction algorithm},
  author={Lorensen, William E and Cline, Harvey E},
  booktitle={Seminal graphics: pioneering efforts that shaped the field},
  pages={347--353},
  year={1998}
}

@inproceedings{park2019deepsdf,
  title={Deepsdf: Learning continuous signed distance functions for shape representation},
  author={Park, Jeong Joon and Florence, Peter and Straub, Julian and Newcombe, Richard and Lovegrove, Steven},
  booktitle={Proceedings of the IEEE/CVF conference on computer vision and pattern recognition},
  pages={165--174},
  year={2019}
}

@inproceedings{kohler2023symmetric,
  title={Symmetric Models for Radar Response Modeling},
  author={Kohler, Colin and Vaska, Nathan and Muthukrishnan, Ramya and Choi, Whangbong and Park, Jung Yeon and Goodwin, Justin and Caceres, Rajmonda and Walters, Robin},
  booktitle={NeurIPS 2023 Workshop on Symmetry and Geometry in Neural Representations},
  year={2023}
}

@article{vaswani2017attention,
  title={Attention is all you need},
  author={Vaswani, A},
  journal={Advances in Neural Information Processing Systems},
  year={2017}
}

@inproceedings{muthukrishnan2023invrt,
  title={InvRT: Solving Radar Inverse Problems with Transformers},
  author={Muthukrishnan, Ramya and Goodwin, Justin and Kern, Adam and Vaska, Nathan and Caceres, Rajmonda S},
  booktitle={AAAI},
  year={2023}
}

@article{ho2020denoising,
  title={Denoising diffusion probabilistic models},
  author={Ho, Jonathan and Jain, Ajay and Abbeel, Pieter},
  journal={Advances in neural information processing systems},
  volume={33},
  pages={6840--6851},
  year={2020}
}

@book{balanis2012advanced,
  title={Advanced engineering electromagnetics},
  author={Balanis, Constantine A},
  year={2012},
  publisher={John Wiley \& Sons}
}
}

% \appendix
% \input{appendices/appendices}

\end{document}